\documentclass{article}
\usepackage{ijcai20}

\usepackage[table]{xcolor}
\usepackage{appendix}
\usepackage{times}
\usepackage{soul}
\usepackage{url}
\usepackage{bm}
\usepackage[hidelinks]{hyperref}
\usepackage[utf8]{inputenc}
\usepackage[small]{caption}
\usepackage{graphicx}
\usepackage{amsmath}
\usepackage{amsthm}
\usepackage{booktabs}
\usepackage{algorithm}
\usepackage{algorithmic}
\usepackage{makecell}
\usepackage{enumitem}
\usepackage{cases}
\usepackage{multirow}
\usepackage{lipsum}
\usepackage{setspace}
\usepackage{graphicx,dblfloatfix,caption,afterpage}

\newcommand{\tabincell}[2]{\begin{tabular}{@{}#1@{}}#2\end{tabular}}

\title {ERNIE-G{\fontsize{13.1pt}{0}\selectfont EN}: An Enhanced Multi-Flow Pre-training and Fine-tuning \\Framework for Natural Language Generation}

\newcommand\blfootnote[1]{%
\begingroup 
\renewcommand\thefootnote{}\footnote{#1}%
\addtocounter{footnote}{-1}%
\endgroup 
}

\author{
Dongling Xiao$^*$\blfootnote{indicates equal contribution.}\and
Han Zhang$^*$\and
Yukun Li\and
Yu Sun\and
Hao Tian\and
\\Hua Wu\And
Haifeng Wang
\affiliations
Baidu Inc., China
\emails
\{xiaodongling, zhanghan17, liyukun01, sunyu02, tianhao, wu\_hua, wanghaifeng\}@baidu.com
}
\begin{document}
\maketitle
\begin{abstract}
Current pre-training works in natural language generation pay little attention to the problem of exposure bias on downstream tasks. To address this issue, we propose an enhanced multi-flow sequence to sequence pre-training and fine-tuning framework named ERNIE-G{\footnotesize EN}, which bridges the discrepancy between training and inference with an infilling generation mechanism and a noise-aware generation method. To make generation closer to human writing patterns, this framework introduces a span-by-span generation flow that trains the model to predict semantically-complete spans consecutively rather than predicting word by word. Unlike existing pre-training methods, ERNIE-G{\footnotesize EN} incorporates multi-granularity target sampling to construct pre-training data, which enhances the correlation between encoder and decoder. Experimental results demonstrate that ERNIE-G{\footnotesize EN} achieves state-of-the-art results with a much smaller amount of pre-training data and parameters on a range of language generation tasks, including abstractive summarization (Gigaword and CNN/DailyMail), question generation (SQuAD), dialogue response generation (Persona-Chat) and generative question answering (CoQA). The source codes and pre-trained models have been released at \url{https://github.com/PaddlePaddle/ERNIE}.
\end{abstract}
\section{Introduction}
Pre-trained on large-scale unlabeled text corpora and fine-tuned on downstream tasks, self-supervised representation models such as GPT \cite{GPT}, BERT \cite{devlin2019bert} and XLNet \cite{XLNet} have achieved remarkable improvements in natural language understanding (NLU). Different from encoder-only pre-training like BERT or decoder-only pre-training like GPT, natural language generation (NLG) relies on the sequence to sequence generation framework (seq2seq) which consists of a bidirectional encoder and a unidirectional decoder. Current pre-training works in NLG such as MASS \cite{MASS} and U{\scriptsize NI}LM \cite{UNILM} mainly focus on jointly pre-training encoder and decoder on different self-supervised tasks. However, these works pay little attention to the exposure bias issue \cite{seq2seq}, a major drawback of teacher-forcing training. This issue is due to the fact that groundtruth words are used during training, while generated words, whether predicted correctly or not, are used for inference where mistakes tend to accumulate. 
\begin{figure}[t]
\centering 
\setlength{\abovecaptionskip}{6pt}
\includegraphics[width=8.5cm]{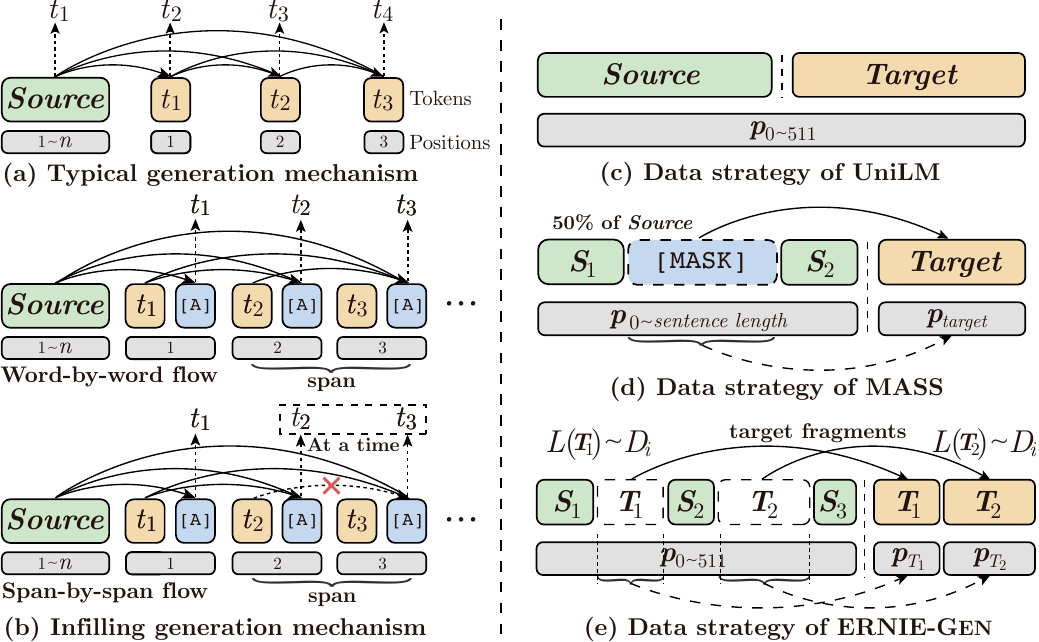}
\caption{Schematic of two generation mechanisms (left) and data strategies for pre-training (right). Blocks in green, orange and blue denote source texts, target texts and artificial symbols.}
\label{fig_1}
\end{figure}
To alleviate this issue, we present ERNIE-G{\footnotesize EN}, an enhanced multi-flow seq2seq training framework characterized by a carefully-designed Multi-Flow Attention architecture based on Transformer \cite{transformer}, as illustrated in Figure \ref{fig_5}. ERNIE-G{\footnotesize EN} incorporates a novel infilling generation mechanism and a noise-aware generation method into pre-training and fine-tuning, which is proved to be effective through experiments in \S\ref{sec_4}. 
\begin{itemize}[leftmargin=*]
\item \textbf{Infilling generation.} 
Instead of using last groundtruth word in training or last generated word in inference, we adopt an inserted artificial symbol {\tt [ATTN]} along with its position to gather history contextual representations at each step in both training and inference, which diverts model's attention away from last word and coerces it into focusing on all former representations, thus alleviating negative influence of previous mistakes to subsequent generation, as shown in Figure \ref{fig_1}(b).
\item \textbf{Noise-Aware generation.} 
We corrupt the input target sequence by randomly replacing words to arbitrary words in the vocabulary. This setup, despite its simplicity, proves to be an effective way to make the model be aware of mistakes in training, so that the model is able to detect mistakes and ignore them during inference.
\end{itemize}

Moreover, in light of the fact that entities, phrases and sentences in human writing are organized in a coherent manner, we incorporate a \textbf{span-by-span generation task} into ERNIE-G{\footnotesize EN} as a new generation flow to train the model to predict semantically-complete spans consecutively rather than predicting word by word as traditional models do. This task is implemented through the infilling generation mechanism in parallel with an infilling-based word-by-word generation flow to facilitate convergence in training, as shown in Figure \ref{fig_1}b. 
\begin{figure}[t]
\centering 
\setlength{\abovecaptionskip}{6pt}
\includegraphics[width=8.4cm]{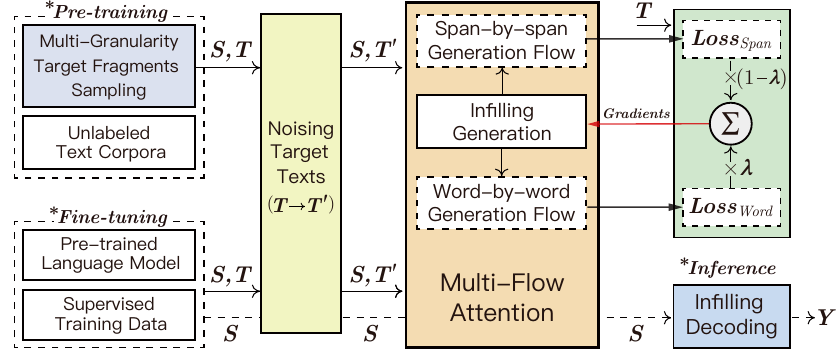}
\caption{Overview of ERNIE-G{\fontsize{8pt}{0}\selectfont EN} framework. $\bm{S},\bm{T}$ and $\bm{Y}$ donate source, target, and generated texts, $\bm{T}^\prime$ is the noised version of $\bm{T}$.}
\label{fig_5}
\end{figure}

In addition, as shown in Figure \ref{fig_1}(c-d), recent pre-training works for NLG like U{\scriptsize NI}LM and MASS only sample a single continuous segment as target sequence. However, this sampling method compromises the correlation between encoder and decoder when it comes to pre-training of long texts (typically 512 words), given that adjacent segments are often relevant semantically. ERNIE-G{\footnotesize EN} adopts a \textbf{multi-granularity target fragments} sampling strategy to force decoder to rely more on the encoder representations other than the previous generated words, thus enhancing the correlation between encoder and decoder, as shown in Figure \ref{fig_1}e. 

Empirically, ERNIE-G{\footnotesize EN} is particularly effective and achieves state-of-the-art results on a range of NLG tasks including abstractive summarization (Gigaword and CN- N/DailyMail), question generation (SQuAD), dialogue response generation (Persona-Chat) and generative question answering (CoQA), utilizing a much smaller amount of pre-training data and parameters.

\section{Related Work}
\paragraph{Pre-training for NLP Tasks.} Recently, pre-training methods have achieved
state-of-the-art results in multiple NLU tasks. ELMo \cite{elmo} pre-trains two unidirectional language models (LMs) with forward and backward direction respectively to feature downstream tasks. GPT utilizes an adjusted Transformer \cite{transformer} to learn a forward LM and then fine-tunes the forward LM on supervised datasets. BERT proposes a masked language modeling (MLM) task to learn deep bidirectional representations. Nevertheless, above methods are usually implemented by just one encoder or decoder, which is less effective in encoder-decoder based generation tasks, thus several works have preliminarily explored the pre-training towards NLG by incorporating BERT's MLM into the seq2seq framework and shown excellent performance on a range of generation tasks. MASS masks a consecutive fragment (50\%) of the input sentence with {\tt [MASK]} symbols to predict. U{\scriptsize NI}LM masks several words in the input sequence which is a pair of segments for encoder and decoder, and then predicts the masked words in accordance with BERT’s MLM. 
\paragraph{Exposure Bias Issue.} NLG tasks suffer from the exposure bias which is caused by teacher-forcing training. To address such issue, RNN-based variational autoencoders (VAEs) are leveraged in \cite{cvae,vae1}, whereas it requires inference for both posterior and prior distribution. Reinforcement learning is also adopted to text generation against exposure bias issue \cite{seq2seq,scst}, which is, however, inefficient during training because of the word-by-word sampling procedure. These methods are inefficient and less practical for pre-training that relies on large-scale unlabeled text corpora.
\paragraph{Span-level Pre-training.} \cite{ernie1,ernie,spanbert} verify that predicting spans reaches substantially better performance on NLU tasks. Meanwhile, inspired by characteristics of human expression, we hope the model have the foresight to generate a semantically-complete span at each step rather than a word. Consequently, a span-by-span generating task is proposed to make the model capable of generating texts more human-like.
\section{Proposed Framework}
 Built on infilling generation mechanism, ERNIE-G{\footnotesize EN} adopts a Multi-Flow Attention architecture to train the model on word-by-word and span-by-span generation tasks in parallel. In this section, we describe ERNIE-G{\footnotesize EN} according to the training process shown in Figure \ref{fig_5}.
\label{sec_3}
\subsection{Multi-Granularity Target Fragments}
 Given an input sequence $\bm{S}=\{s_{1},...,s_{n}\}$, we first sample a length distribution $D_{i}$ from a distribution set $\bm{D}=\{D_{1},...,D_{|\bm{D}|}\}$ with probability $p_{i}$ for target fragments, and then select fragments according to $D_{i}$ in  $\bm{S}$ iteratively until the fragment budget has been spent (e.g. 25\% of $\bm{S}$). We denote $S_j^{i}$ as the $j$-th fragment which is sampled in length distribution $D_{i}$. Sampled fragments are then removed from $\bm{S}$ and stitched together to form target sequence $\bm{T}=[T_{1},...,T_{k}]=[S_1^{i},...,S_k^{i}]$. We denote $\bm{S}^\prime$ as the left input sequence after removing sampled fragments.
 \begin{figure*}[t]
	\centering
	\setlength{\abovecaptionskip}{5pt}
	\includegraphics[width=17.4cm]{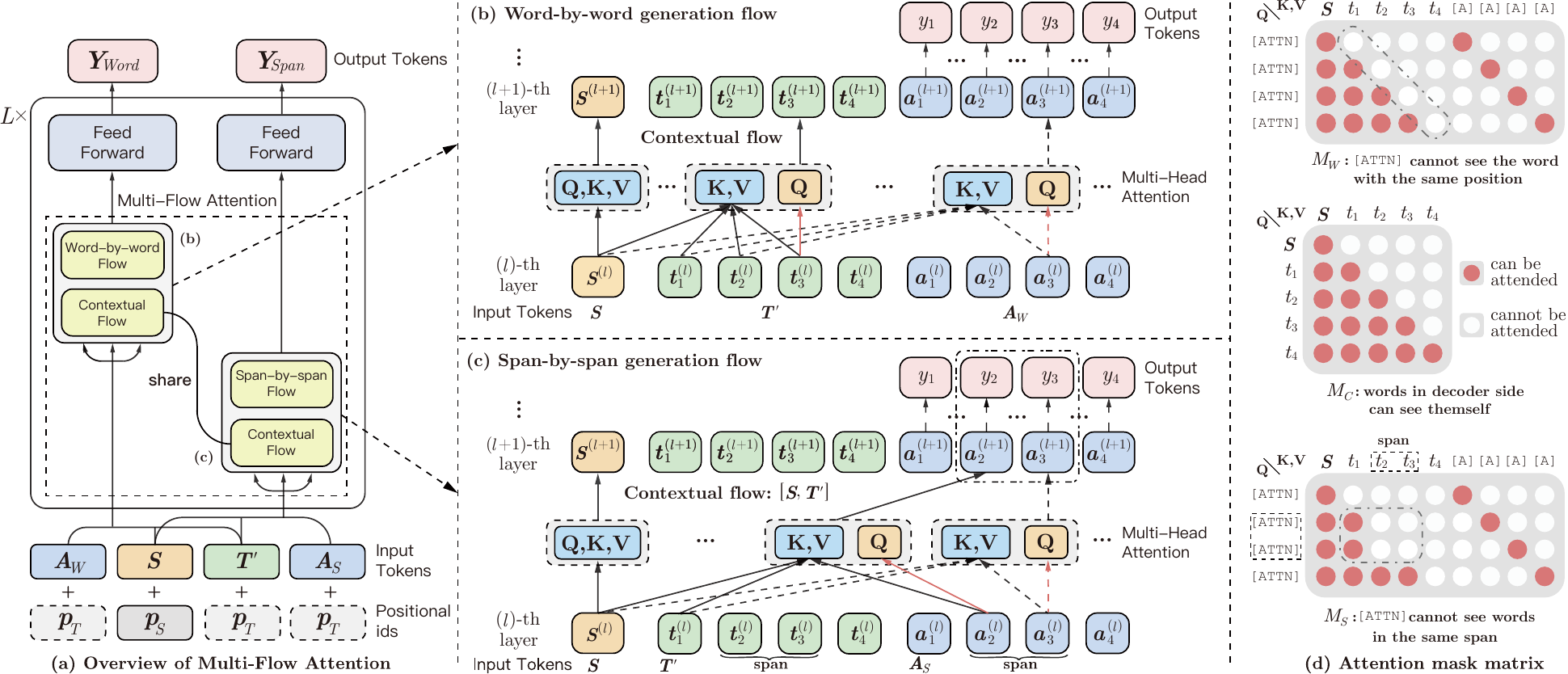}
	\caption{
	Illustration of the Multi-Flow Attention module. (a):Overview of multi-flow attention. The encoder and the decoder share the parameters of multi-layer Transformer. (b):Word-by-word generation flow with history contextual representations from \textit{Contextual Flow}. (c):Span-by-span generation flow with shared \textit{Contextual Flow}. (d):The attention mask matrixes of word-by-word generation flow ($M_{W}$), contextual flow ($M_{C}$) and span-by-span generation flow ($M_{S}$). 
	The $i$-th generated token $y_i$ is calculated by $\textstyle{\texttt{argmax}(\texttt{softmax}(\texttt{Fc}(\bm{a}_i^{(L-1)})))}$. 
	}
	\label{fig_2}
\end{figure*}
ERNIE-G{\footnotesize EN} performs pre-training by predicting the fragmented target sequence $\bm{T}$ and minimizing the negative log likelihood:
\begin{equation}
 	\begin{split}
 	\mathcal{L}(\theta;\bm{S},D_{i})&=-{\rm log}P(\bm{T}|\bm{S}^\prime,D_{i};\theta)   \\[-0.3mm]
  	&=-{\rm log}\prod_{j=1}^k P(T_{j}|T_{<j},\bm{S}^\prime,D_{i};\theta).
    \end{split}
\end{equation}
where the target sequence $\bm{T}$ is sorted by the positions of sampled fragments. For each fragment $T=\{t_{1},...,t_{|T|}\}$ in $\bm{T}$, we have $P(T)=\prod_{j=1}^{|T|}P(t_j|t_{<j})$.

Following preliminary trials, we set a hyperparameter $\gamma=0.25$, 
which denotes the ratio of length of all fragments to that of the input sequence $\bm{S}$. Besides, we introduce two uniform distributions $\bm{D}=\{U(1,4), U(4,32)\}$ with probability of $0.4$ and $0.6$ respectively to sample fragments, which aims to learn representations from different perspectives. On the one hand, short fragments benefit learning of semantic relation between words; on the other hand, longer fragments help to memorize sentence-level expressions.
\subsection{Noise-Aware Generation}

To train a generation model which can detect the false prediction and mitigate its impact on subsequent generation, we corrupt the groundtruth sequence
 $\bm{T}$ with a procedure where words are being replaced randomly, and the corrupted $\bm{T}$ is represented as $\bm{T}^{\prime}$. There are two hyperparameters, $\rho_{p}$ and $\rho_{f}$, denoting the noising rate in pre-training and fine-tuning respectively.
\subsection{Architecture: Multi-Flow Attention}
Formally, given a source sequence $\bm{S}=\{s_{1},...,s_{n}\}$, a noised target sequence $\bm{T}^\prime=\{t_{1},...,t_{m}\}$, we denote the inference of seq2seq network based on shared Transformer as follows: 
\begin{equation}
    \setlength{\abovedisplayskip}{3pt}
    \setlength{\belowdisplayskip}{1pt}
 	\begin{split}
 	\bm{s}_i^{(l+1)}&\leftarrow\texttt{MH-Attn}(Q=\bm{s}_i^{(l)},KV=\bm{S}^{(l)}).  \\
  	\bm{t}_i^{(l+1)}&\leftarrow\texttt{MH-Attn}(Q=\bm{t}_i^{(l)},KV=\left[\bm{S}^{(l)},\bm{t}_{\leq i}^{(l)}\right]).
    \end{split}
\end{equation}
where $Q$, $K$, $V$ denote the query, key, and value in Multi-Head attention \cite{transformer}. $\displaystyle{\bm{s}_i^{(l)}}$ and $\bm{t}_i^{(l)}$ indicate the $i$-th vector representations of the $l$-th layer of Multi-Head Attention for the encoder and the decoder respectively, $[\cdot]$ denotes the concatenation operation. In this work, we call the above procedure the \textbf{\textit{Contextual Flow}}.
\paragraph{Word-by-word Generation Flow.} Based on infilling generation mechanism, this flow utilizes an inserted {\tt [ATTN]} symbol to gather history representations word by word (see Figure 1b). To facilitate this process, we place all inserted {\tt [ATTN]} together to construct an artificial symbol sequence $\bm{A}_W\!=\!\{${\tt [ATTN]}$_{1},...,${\tt [ATTN]}$_{m}\}$ which has the same length as $\bm{T}^\prime$, as shown in Figure \ref{fig_2}b. To be specific, the word-by-word generation flow is updated as follow:
\begin{equation}
\setlength{\abovedisplayskip}{4pt}
\setlength{\belowdisplayskip}{1pt}
\resizebox{.91\linewidth}{!}{$
    \displaystyle
    \bm{a}_i^{(l+1)}\leftarrow\texttt{MH-Attn}(Q=\bm{a}_i^{(l)},KV=\left[\bm{S}^{(l)},\bm{t}_{<i}^{(l)},\bm{a}_i^{(l)}\right]).
$}
\end{equation}
where $\bm{a}_i^{(l)}$ indicates the $i$-th vector representation of the $l$-th layer for the artificial symbol sequence $\bm{A}_W$.
\paragraph{Span-by-span Generation Flow.} Different from word-by-word generation flow, span-by-span flow uses {\tt [ATTN]} symbols to predict spans consecutively, as shown in Figure \ref{fig_2}c. Formally, given a list of span boundaries $B\!=\!\{b_{1},...,b_{|B|}\}$, we conduct the span-by-span generation flow as:
\begin{equation}
\setlength{\abovedisplayskip}{4pt}
\setlength{\belowdisplayskip}{1pt}
\resizebox{.91\linewidth}{!}{$
    \displaystyle
    \bm{a}_{j}^{(l+1)}\leftarrow\texttt{MH-Attn}(Q=\bm{a}_{j}^{(l)},KV=\left[\bm{S}^{(l)},\bm{t}_{<b_{i}}^{(l)},\bm{a}_j^{(l)}\right]).
$}
\end{equation}
where $j\in[b_{i},b_{i+1})$, and $\bm{a}_{j}^{(l)}$ denotes the $(j-b_{i})$-th vector representation of the $i$-th span. Essentially, the model is trained to predict a whole span $\{t_{b_{i}},...,t_{b_{i+1}-1}\}$ with the same history context $\left[\bm{S},\bm{t}_{<b_{i}}\right]$.
Instead of randomly sampling spans, we prefer sampling spans with semantical information and knowledge. Specifically, we consider the following two steps to sample spans consecutively in $\bm{T}^\prime$:
\begin{itemize}[leftmargin=*]
\item Firstly, we implement a T-test to compute t-statistic scores of all bigrams and trigrams, which is based on an initial hypothesis $H_{0}$: a random span of $n$ arbitrary words $\bm{w}\!=\!\{w_{1},...,w_{n}\}$ with probability $p^\prime (\bm{w})\!=\!\prod_{i=1}^n p(w_{i})$ cannot be a statistical $n$-gram. The t-statistic score is calculated by ${\frac{(p(\bm{w})-p^\prime(\bm{w}))}{\sqrt{\sigma^2/N}}}$, where $p(\bm{w})\!=\!\frac{\texttt{Count}(\bm{w})}{N}$ and $\sigma^2\!\!=\!p(\bm{w})(1-p(\bm{w}))$, indicating the statistic probability and the standard deviation of $\bm{w}$ respectively, $N$ denotes the total number of $n$-grams appearing in the training data. According to the t-statistic scores, we select the top 200,000 bigrams, top 50,000 trigrams and all unigrams to construct a specific vocabulary of spans, which is represented as $\bm{V}_{span}$. 
\item Secondly, we search the trigram, bigram and unigram in order, starting with current word until a span ($n$-gram, $n\leq3$) is retrieved in $\bm{V}_{span}$.
\end{itemize}
\paragraph{Multi-Flow Attention.} To integrate the word-by-word generation flow and span-by-span generation flow, we apply them in parallel with a shared contextual flow by leveraging the multi-flow attention architecture, as Figure \ref{fig_2}a describes. The multi-flow attention is computed as:
\begin{equation}
\setlength{\abovedisplayskip}{3pt}
\setlength{\belowdisplayskip}{2pt}
~~\begin{cases}\resizebox{.73\linewidth}{!}{$\bm{X}^{(l+1)}\!\leftarrow\!\texttt{MH-Attn}(Q\!=\!\bm{X}^{(l)},KV\!=\!\bm{X}^{(l)},M_{C})$} \\[1.6mm]
\resizebox{.8\linewidth}{!}{$\bm{A}_W^{(l+1)}\!\leftarrow\!\texttt{MH-Attn}(Q\!=\!\bm{A}_W^{(l)},KV\!=\!\left[\bm{X}^{(l)}\!,\bm{A}_W^{(l)}\right]\!,M_{W})$}\! \\[1.8mm]
\resizebox{.8\linewidth}{!}{$\bm{A}_S^{(l+1)}\!\leftarrow\!\texttt{MH-Attn}(Q\!=\!\bm{A}_S^{(l)},KV\!=\!\left[\bm{X}^{(l)}\!,\bm{A}_S^{(l)}\right]\!,M_{S})$}\!
\end{cases}
\end{equation}
where $
\bm{X}$ denotes the concatenation of $\bm{S}$ and $\bm{T}^{\prime}$, $\bm{X}^{(l)}$ is the vector sequence of the $l$-th layer for the contextual flow. $\bm{A}_W^{(l)}$, $\bm{A}_S^{(l)}$ are vector sequences of the $l$-th layer for the word-by-word and span-by-span generation flow respectively. As shown in Figure \ref{fig_2}d, attention mask matrix $M$ determines whether query and key can attend to each other by modifying the attention weight $W\!\!=\!\texttt{softmax}(\frac{QK^T}{\sqrt{d_k}}+M)$ \cite{transformer} . Specifically, $M$ is assigned as:
\begin{equation}
\setlength{\abovedisplayskip}{3pt}
\setlength{\belowdisplayskip}{3pt}
M_{ij}=\begin{cases}
0,~~~~~~~ {\rm can~be~attended}\\
-\infty,~~{\rm prevent~from~attending}
\end{cases}
\end{equation}

While training, we add the loss of the word-by-word and span-by-span generation flow with an coefficient $\lambda$:
\begin{equation}
\setlength{\abovedisplayskip}{5pt}
\setlength{\belowdisplayskip}{3pt}
\resizebox{.91\linewidth}{!}{$
 	\begin{split}
 	\mathcal{L}(\bm{T})&=\lambda\mathcal{L}_{Word}(\bm{T})+(1-\lambda)\mathcal{L}_{Span}(\bm{T})
 	\\
 	&=-\lambda{\rm log}P(\bm{T}|\bm{A}_W^{(L-1)})-(1-\lambda){\rm log}P(\bm{T}|\bm{A}_S^{(L-1)}). 
    \end{split}
$}
\end{equation}
where $\bm{T}$ indicates the unnoised target sequence, and $\mathcal{L}(\cdot)$ denotes the cross entropy loss function. In detail, we set $\lambda=0.5$ and $\lambda=1.0$ respectively in pre-training and fine-tuning.
\subsection{Inference: Infilling Decoding}
\begin{figure}[t]
\setlength{\belowcaptionskip}{-0.5cm} 
\setlength{\abovecaptionskip}{5pt}
\centering
\includegraphics[width=6.7cm]{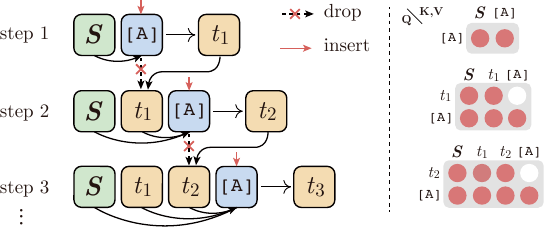}
\caption{Schematic of infilling decoding: the particular procedures in infilling decoding including dropping and inserting (left) and the attention mask matrixes at each step (right).}
\label{fig_3}
\end{figure}

During inference, the target sequence $\bm{T}$ is unknown, we insert symbol {\tt [ATTN]} step by step to gather the representation of history context instead of preparing an artificial symbol sequence $\bm{A}$ in advance. Meanwhile, for the purpose of efficiency, we need to drop the inserted {\tt [ATTN]} after inference at each step, as detailed in Figure \ref{fig_3}. 
\section{Experiments}
In this section, we compare our ERNIE-G{\footnotesize EN} with previous works and conduct several ablation experiments to assess the performance of proposed methods in \S\ref{sec_3}. 
\label{sec_4}
\subsection{Pre-training and Implementation}
Analogous to BERT and U{\scriptsize NI}LM, ERNIE-G{\footnotesize EN} is trained on English Wikipedia and BookCorpus \cite{book}, totaling 16GB. We also pre-train ERNIE-G{\footnotesize EN} on larger scaled text corpora, which is specifically described in appendix \ref{app}. The input sequence is lowercased and truncated to a maximum length of 512. We train a base model ERNIE-G{\footnotesize EN}$_{BASE}$ ($L$=12, $H$=768, $A$=12, Total Parameters=110M)\footnote{We donate the number of layers as $L$, the hidden size as $H$ and the number of self-attention heads as $A$.} and a large model ERNIE-G{\footnotesize EN}$_{LARGE}$ ($L$=24, $H$=1024, $A$=16, Total Parameters=340M) with parameters initialized by BERT$_{BASE}$ and BERT$_{LARGE}$ respectively. Specifically, Adam optimizer with $\beta_1=0.9,\beta_2=0.999,\epsilon=10^{-9}$ is employed. The peak learning rate is 5e-5 with warmup over the first 4,000 steps and linear decay scheduling. The noising rate $\rho_{p}$ for pre-training is 0.05. Batches are organized by limiting the maximum number of tokens to 196,608. Pre-training experiments are carried out on PaddlePaddle platforms\footnote{\url{https://github.com/PaddlePaddle/Paddle}} and Nvidia Tesla V100 GPU. By virtue of float16 mixed precision training, it takes almost 4 days for 400,000 steps to train ERNIE-G{\footnotesize EN}$_{BASE}$ while almost 7 days for 450,000 steps to train ERNIE-G{\footnotesize EN}$_{LARGE}$.
\subsection{Fine-tuning on Downstream Tasks}
\begin{table*}[t]
\centering
\setlength{\abovecaptionskip}{5pt}
\setlength{\belowcaptionskip}{-0.35cm}
\resizebox{0.91\textwidth}{!}{
\begin{tabular}{c|ccccccccccccc}  
\toprule[1.0pt]
\multirow{2}{*}{Task} &\multicolumn{2}{c}{Epoch}& \multicolumn{2}{c}{Learning Rate} & \multicolumn{2}{c}{Noising Rate $\rho_{f}$}&\multicolumn{2}{c}{Dropout Rate}& Batch & Label&Beam&\multirow{2}{*}{Evaluation Metric} \\
&$\scriptstyle{BASE}$&$\scriptstyle{LARGE}$&$\scriptstyle{BASE}$&$\scriptstyle{LARGE}$&$\scriptstyle{BASE}$&$\scriptstyle{LARGE}$&$\scriptstyle{BASE}$&$\scriptstyle{LARGE}$&Size&Smooth&Size&&\\
\midrule[0.3pt]
SQuAD QG &10&10&2.5e-5&1.5e-5&0.7&0.7&0.1&0.2&32&0.1&1& BLEU-4, METEOR (MTR), ROUGE-L (RG-L) \\
\midrule[0.3pt]
CNN/DailyMail &30&20&5e-5&4e-5 & 0.7&0.7 &0.1&0.1&64&0.1&5&\multirow{2}{*}{\tabincell{c}{ROUGE-F1 scores:\\ ROUGE-1 (RG-1), ROUGE-2 (RG-2), ROUGE-L (RG-L)}}\\
Gigaword &10&5&3e-5&3e-5 & 0.5&0.6 &0.1&0.2&128&0.1&5&\\
\midrule[0.3pt]
Persona-Chat &-&30&-&1e-4 & -&0.0 &-&0.1&64&0.1&10&BLEU-1, BLEU-2, Distinct-1, Distinct-2\\
\midrule[0.5pt]
Generative CoQA &-&10&-&1e-5&-&0.5 &-&0.1&32&0.1&3&F1-score\\
\bottomrule[1.0pt]
\end{tabular}
}
\caption{Hyperparamters of fine-tuning for ERNIE-G{\fontsize{8pt}{0}\selectfont EN}$_{BASE}$ and ERNIE-G{\fontsize{8pt}{0}\selectfont EN}$_{LARGE}$.}
\label{tab_1}
\end{table*}
\paragraph{Abstractive Summarization} aims at generating fluent and concise summaries without being constrained to extracting sub-sequences from the input articles. We execute experiments on Gigaword dataset \cite{gigaword} and CNN/D-ailyMail dataset \cite{cnn}. Gigaword dataset contains 3.8M articles extracted from the Gigaword corpus, while CNN/DailyMail dataset consists of 93k articles and 220k articles from the CNN and Daily Mail respectively. 
\begin{table}[H]
\centering
\addtolength{\tabcolsep}{-1mm}
\resizebox{0.485\textwidth}{!}{
\begin{tabular}{l|cc|c}  
\toprule[1.0pt]
\textbf{Model} & \textbf{{\footnotesize Data}} & \textbf{{\footnotesize Params}} & \textbf{{\footnotesize RG-1 / RG-2 / RG-L}}  \\
\midrule[0.3pt]
\multicolumn{4}{l}{\textit{* 10k training samples : Gigaword 10k}} \\
MASS \cite{MASS}&18G&160M&25.03 / 9.48 / 23.48 \\
U{\scriptsize NI}LM$_{LARGE}$ \cite{UNILM}&16G&340M&32.96 / 14.68 / 30.56\\
\midrule[0.4pt]
ERNIE-G{\footnotesize EN}$_{BASE}$&16G&110M&33.75 / 15.23 / 31.35\\
ERNIE-G{\footnotesize EN}$_{LARGE}$&16G&340M&\textbf{35.05} / \textbf{16.10} / \textbf{32.50}\\
\midrule[0.6pt]
\multicolumn{4}{l}{\textit{* Fully 3.8M training samples}} \\
MASS \cite{MASS}&18G&160M&37.66 / 18.53 / 34.89\\
{\footnotesize BERT}S{\footnotesize HARE} \cite{bertshare}&16G&110M&38.13 / 19.81 / 35.62\\
U{\scriptsize NI}LM$_{LARGE}$ \cite{UNILM}&16G&340M&38.45 / 19.45 / 35.75\\
PEGASUS{\scriptsize (C4)} \cite{pegasus}&750G&568M&38.75 / 19.96 / 36.14\\
PEGASUS{\scriptsize (HugeNews)} \cite{pegasus}&3.8T&568M&39.12 / 19.86 / 36.24\\
\midrule[0.4pt]
ERNIE-G{\footnotesize EN}$_{BASE}$&16G&110M&38.83 / 20.04 / 36.20\\
ERNIE-G{\footnotesize EN}$_{LARGE}$&16G&340M&\textbf{39.25} / \textbf{20.25} / \textbf{36.53}\\
\bottomrule[1.0pt]	
\end{tabular}
}
\caption{Comparison on Gigaword dataset with state-of-the-art results. Models in the upper block use 10k sample for fine-tuning. We also report the size of pre-training data and parameters utilized for each listed model (columns 2-3). RG is short for ROUGE.}
\label{tab_2}
\end{table}
The results on Gigaword task with two scales (10k and 3.8M) are presented in Table \ref{tab_2}, and the fine-tuning settings are shown in Table \ref{tab_1}. On the low-resource task (Gigaword 10k), ERNIE-G{\footnotesize EN}$_{LARGE}$ yields a gain of $+1.94$ ROUGE-L compared with U{\scriptsize NI}LM$_{LARGE}$. On the full training set, ERNIE-G{\footnotesize EN}$_{LARGE}$ creates the state-of-the-art results, outperforming various previous methods. Specifically, ERNIE-G{\footnotesize EN}$_{BASE}$ outperforms PEGASUS (568M and 750G) by using only 110M parameters and 16G training data.
\begin{table}
\centering
\setlength{\abovecaptionskip}{5pt}
\setlength{\belowcaptionskip}{-0.3cm}
\addtolength{\tabcolsep}{-1mm}
\resizebox{0.485\textwidth}{!}{
\begin{tabular}{l|cc|c}
\toprule[1.0pt]
\textbf{Model} & \textbf{{\footnotesize Data}} & \textbf{{\footnotesize Params}} & \textbf{{\footnotesize RG-1 / RG-2 / RG-L}}  \\
\midrule[0.3pt]
{\footnotesize BERT}S{\footnotesize HARE} \cite{bertshare}&16G&110M&39.25 / 18.09 / 36.45\\
B{\footnotesize ERT}S{\footnotesize UM}A{\footnotesize BS} \cite{bertsum}&16G&110M&41.72 / 19.39 / 38.76\\
MASS \cite{MASS}&18G&160M&42.12 / 19.50 / 39.01\\
U{\scriptsize NI}LM$_{LARGE}$ \cite{UNILM}&16G&340M&43.33 / 20.21 / 40.51\\
T5$_{LARGE}$ \cite{T5}&750G&340M&42.50 / 20.68 / 39.75\\
T5$_{XLARGE}$ \cite{T5}&750G&11B&43.52 / \textbf{21.55} / 40.69\\
BART$_{LARGE}$ \cite{BART}&430G&400M&44.16 / 21.28 / 40.90\\
PEGASUS{\scriptsize (C4)} \cite{pegasus}&750G&568M&43.90 / 21.20 / 40.76\\
PEGASUS{\scriptsize (HugeNews)} \cite{pegasus}&3.8T&568M&\textbf{44.17} / 21.47 / 41.11\\
\midrule[0.3pt]
ERNIE-G{\footnotesize EN}$_{BASE}$&16G&110M&42.30 / 19.92 / 39.68\\
ERNIE-G{\footnotesize EN}$_{LARGE}$&16G&340M&44.02 / 21.17 / \textbf{41.26}\\
\bottomrule[1.0pt]	
\end{tabular}
}
\caption{Evaluation results on CNN/DailyMail. C4 and HugeNews are two massive datasets of 750G and 3.8T respectively.}
\label{tab_3}
\end{table}

Table \ref{tab_3} shows the performance on CNN/DailyMail. With a similar amount of pre-training data and parameters, ERNIE-G{\footnotesize EN}$_{BASE}$ outperforms MASS by $+0.67$ ROUGE-L scores. Fairly compared with U{\scriptsize NI}LM$_{LARGE}$, ERNIE-G{\footnotesize EN}$_{LARGE}$ obtains substantial gain of $+0.73$ ROUGE-L scores. Meanwhile, in spite of small pre-training data and parameters, our large model also achieves state-of-the-art result on ROUGE-L and comparable performance on ROUGE-1/2.
\begin{table}[b]
\centering
\setlength{\abovecaptionskip}{8pt}
\addtolength{\tabcolsep}{-1mm}
\resizebox{0.45\textwidth}{!}{
\begin{tabular}{l|ccc}  
\toprule[1.0pt]
\textbf{Model} & \textbf{{\footnotesize BLEU-4}}~ & \textbf{{\footnotesize METEOR}} & \textbf{{\footnotesize ROUGE-L}}  \\
\midrule[0.3pt]
SemQG \cite{semiq}&20.76&24.20&48.91\\
U{\scriptsize NI}LM$_{LARGE}$ \cite{UNILM}&22.12&25.06&51.07\\
\midrule[0.3pt]
ERNIE-G{\footnotesize EN}$_{BASE}$\ \ (beam size = $1$)&22.28&25.13&50.58\\
ERNIE-G{\footnotesize EN}$_{LARGE}$ (beam size = $1$)&24.03&26.31&52.36\\
ERNIE-G{\footnotesize EN}$_{LARGE}$ (beam size = $5$)&\textbf{25.40}&\textbf{26.92}&\textbf{52.84}\\
\midrule[0.6pt]
\multicolumn{4}{l}{* \textit{Reversed test} $\leftrightarrow$ \textit{dev split}} \\
MP-GSN \cite{split}&16.38&20.25&44.48\\
SemQG \cite{semiq}&20.76&24.20&48.91\\
U{\scriptsize NI}LM$_{LARGE}$ \cite{UNILM}&23.75&25.61&52.04\\
\midrule[0.3pt]
ERNIE-G{\footnotesize EN}$_{BASE}$\ \ (beam size = $1$)&23.52&25.61&51.45\\
ERNIE-G{\footnotesize EN}$_{LARGE}$ (beam size = $1$)&25.57&26.89&53.31\\
ERNIE-G{\footnotesize EN}$_{LARGE}$ (beam size = $5$)&\textbf{26.95}&\textbf{27.57}&\textbf{53.77}\\
\bottomrule[1.0pt]	
\end{tabular}
}
\caption{Question generation results on SQuAD. Models in the upper block and the lower block use different \textit{test} $\leftrightarrow$ \textit{dev} split method.}
\label{tab_4}
\end{table}
\paragraph{Question Generation} is to generate a question according to a given input passage and a corresponding answer. We evaluate on the SQuAD 1.1 dataset \cite{squad} for question generation task (called SQuAD QG). Following U{\scriptsize NI}LM, we redistribute the original dataset into a new training set and testing set with the original development set unchanged. We also conduct experiment with the reversed dev$\leftrightarrow$test split as \cite{split} indicates. In Table \ref{tab_4}, we present the results of ERNIE-G{\footnotesize EN} and several previous works. Again, ERNIE-G{\footnotesize EN} outperforms U{\scriptsize NI}LM$_{LARGE}$ and achieves a new state-of-the-art result on question generation by giving  $+1.82$ BLEU-4 scores.
\begin{table}
\centering
\setlength{\abovecaptionskip}{5pt}
\setlength{\belowcaptionskip}{-0.3cm}
\resizebox{0.433\textwidth}{!}{
\begin{tabular}{l|cc}  
\toprule[1.0pt]
\textbf{Model} & \textbf{{\footnotesize BLEU-1/2}} & \textbf{{\footnotesize Distinct-1/2}}\\
\midrule[0.3pt]
LIC \cite{plato}&40.5 / 32.0&0.019 / 0.113 \\
PLATO$_{{\rm w/o~latent}}$ \cite{plato}&45.8 / 35.7&0.012 / 0.064\\
PLATO \cite{plato}&40.6 / 31.5&0.021 / 0.121\\
\midrule[0.3pt]
ERNIE-G{\footnotesize EN}$_{LARGE}$&\textbf{46.8} / \textbf{36.4}&\textbf{0.023} / \textbf{0.168}\\
\bottomrule[1.0pt]	
\end{tabular}
}
\caption{Comparison with state-of-the-art results on Persona-Chat.}
\label{tab_5}
\end{table}
\paragraph{Generative Question Answering / Dialogue Response} in multi-turn conversations are challenging because of complex background knowledge and diverse utterances. We conduct an experiment on Persona-Chat dataset \cite{persona} to generate responses according to given multi-turn conversations and persona profile. Table \ref{tab_5} shows that ERNIE-G{\footnotesize EN} outperforms current task-specific pre-training model on dialogue generation. Beside, we also execute an experiment on CoQA dataset \cite{coqa} to generate free-form answers for input questions and conversations. As shown in Table \ref{tab_6}, our generative question answering model works considerably better than early works by $+2.0$ F1-scores.

\begin{table}[H]
\centering
\setlength{\abovecaptionskip}{6pt}
\resizebox{0.315\textwidth}{!}{
\begin{tabular}{l|c}  
\toprule[1.0pt]
\textbf{Model} & \textbf{{\footnotesize F1-score}}\\
\midrule[0.3pt]
Seq2Seq \cite{coqa}&27.5\\
PGNet \cite{coqa}&45.4\\
U{\scriptsize NI}LM$_{LARGE}$ \cite{UNILM}&82.5\\
\midrule[0.3pt]
ERNIE-G{\footnotesize EN}$_{LARGE}$&\textbf{84.5}\\
\bottomrule[1.0pt]	
\end{tabular}
}
\caption{Generative question answering results on the development set of CoQA.}
\label{tab_6}
\end{table}

\subsection{Ablation Studies}
\label{sec_4}
To better understand the importance of each proposed generation methods, we conduct experiments concerning the following two aspects:

\begin{figure*}[t]
	\centering
	\setlength{\abovecaptionskip}{2pt}
    \setlength{\belowcaptionskip}{-0.2cm}
	\includegraphics[width=16.6cm]{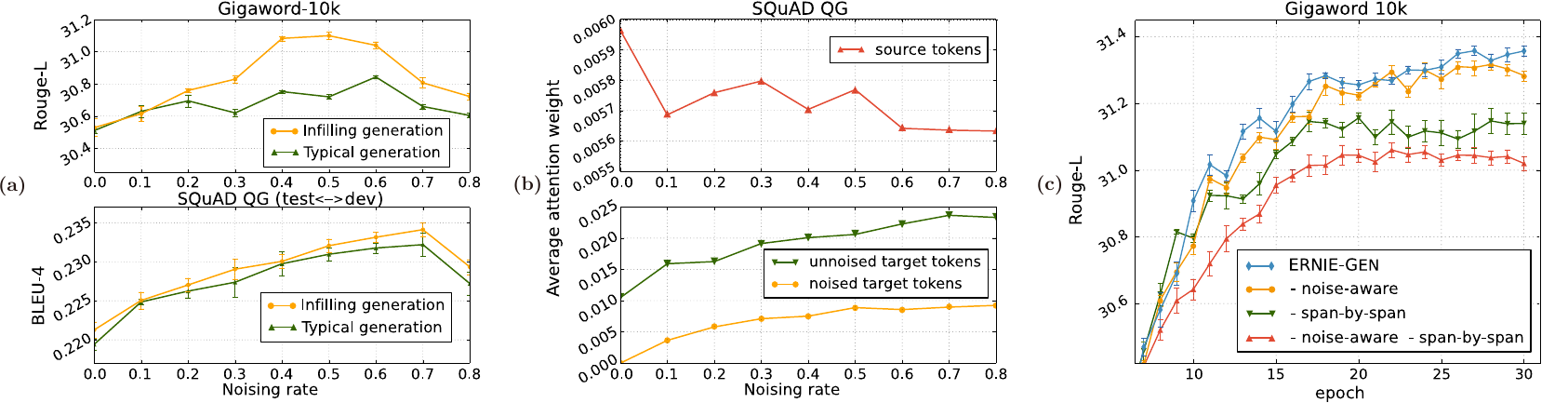}
	\caption{Results of ablation studies. (a):Comparisons between typical generation and infilling generation on Gigaword 10k and SQuAD QG with different fine-tuning noising rate $\rho_{f}$. (b):Noising Analysis, average attention weights of source words, unnoised target words and noised target words for diverse fine-tuning noising rate $\rho_{f}$. (c):Ablation study on Gigaword 10k, the x-axis shows fine-tuning epochs.}
	\label{fig_4}
\end{figure*}

\begin{table*}[t]
\centering
\setlength{\abovecaptionskip}{6pt}
\setlength{\belowcaptionskip}{-0.2cm}
\addtolength{\tabcolsep}{-1mm}
\renewcommand{\arraystretch}{1.05}
\resizebox{1.0\textwidth}{!}{
\begin{tabular}{l|ccc||ccc}
\toprule[1.0pt]
\textbf{{\fontsize{8.5pt}{0}\selectfont \# Fine-tuning method}}&\multicolumn{3}{c||}{1 \textit{Noising fine-tuning}: {\fontsize{8pt}{0}\selectfont \textit{Fine-tuning with noise-aware generation}}} &\multicolumn{3}{c}{2 {\textit{Masking fine-tuning}: {\fontsize{8pt}{0}\selectfont \textit{Only updating the gradients of masked words}}}}\\
\midrule[0.3pt]
\multirow{2}{*}{\#\ \ \textbf{Model}}& \textbf{Gigaword 10k} & \textbf{CNN/DailyMail 10k} & \textbf{SQuAD QG} & \textbf{Gigaword 10k} & \textbf{CNN/DailyMail 10k} & \textbf{SQuAD QG}  \\
& RG-1\,/\,RG-2\,/\,RG-L & RG-1\,/\,RG-2\,/\,RG-L & Bleu-4\,/\,MTR\,/\,RG-L&RG-1\,/\,RG-2\,/\,RG-L & RG-1\,/\,RG-2\,/\,RG-L & Bleu-4\,/\,MTR\,/\,RG-L\\
\midrule[0.3pt]
1\ \ ERNIE-G{\footnotesize EN} & \textbf{33.75}\,/\,\textbf{15.23}\,/\,\textbf{31.35}&\textbf{39.92}\,/\,17.46\,/\,\textbf{37.40}&\textbf{23.52}\,/\,\textbf{25.61}\,/\,\textbf{51.45}& \textbf{33.30}\,/\,\textbf{15.04}\,/ \textbf{31.22}  & \textbf{39.54}\,/\,\textbf{17.61}\,/\,\textbf{37.00} & 22.99\,/\,25.14\,/\,51.31\\
2\ \ \ \ - noise-aware & 33.57\,/\,15.15\,/\,31.28  & 39.78\,/\,\textbf{17.63}\,/\,37.23 & 23.40\,/\,25.50\,/\,51.36& 33.01\,/\,14.94\,/\,31.00  & 39.53\,/\,\textbf{17.61}\,/\,36.97 & 23.09\,/\,\textbf{25.15}\,/\,51.41 \\
3\ \ \ \ - span-by-span & 33.43\,/\,15.04\,/\,31.14  & 39.75\,/\,17.62\,/\,37.21 & 23.37\,/\,25.56\,/\,51.32& 32.97\,/\,14.92\,/\,30.94 & \textbf{39.54}\,/\,17.57\,/\,36.95  & \textbf{23.10}\,/\,25.14\,/\,\textbf{51.42} \\
4\ \ \ \ - 2 and 3 & 33.23\,/\,14.77\,/\,31.00 & 39.71\,/\,17.55\,/\,37.18 & 23.34\,/\,25.54\,/\,51.30& 32.57\,/\,14.68\,/\,30.60 & 39.49\,/\,17.66\,/\,36.96 & 22.89\,/\,25.08\,/\,51.28 \\
\bottomrule[1.0pt]	
\end{tabular}
}
\caption{Ablation study for ERNIE-G{\fontsize{8pt}{0}\selectfont  EN}$_{BASE}$ and its variants. Particularly, We set $\rho_{p}\!=\!0.05$ in pre-training (row 1), while  removing the span-by-span generation task (row 3), we set $\rho_{p}\!=\!0.2$ because the pre-training becomes easier.}
\label{tab_7}
\end{table*}

\begin{itemize}[leftmargin=*]
	\item The robustness of infilling generation mechanism and noise-aware generation method against the exposure bias. 
	\item The effectiveness of span-by-span generation task and the complete ERNIE-G{\footnotesize EN} model.
\end{itemize}

\begin{table}
\centering
\setlength{\abovecaptionskip}{8pt}
\setlength{\belowcaptionskip}{-0.3cm}
\addtolength{\tabcolsep}{-1.3mm}
\resizebox{0.48\textwidth}{!}{
\begin{tabular}{ll|cc}  
\toprule[1.0pt]
\#&Task {\fontsize{7pt}{0}\selectfont (Metrics)}& Typical generation & Infilling generation  \\
\midrule[0.3pt]
\multicolumn{4}{l}{\textit{Fine-tuning without noise-aware generation}} \\
1&Gigaword 10k {\fontsize{7pt}{0}\selectfont (RG-1 / RG-2 / RG-L)} & \textbf{32.98} / \textbf{14.67} / 30.51 & 32.93 /14.46 / \textbf{30.53}\\
2&CNN/DM 10k {\fontsize{7pt}{0}\selectfont (RG-1 / RG-2 / RG-L)}& 39.25 / 16.70 / 36.65 & \textbf{39.56} / \textbf{16.93} / \textbf{36.94}\\
3&SQuAD QG {\fontsize{7pt}{0}\selectfont (Bleu-4 / MTR / RG-L)}& 21.95 / 24.53 / 50.34 & \textbf{22.13} / \textbf{24.66} / \textbf{50.51}\\
\midrule[0.3pt]
\multicolumn{3}{l}{\textit{Fine-tuning with noise-aware generation}} \\
4&Gigaword 10k {\fontsize{7pt}{0}\selectfont (RG-1 / RG-2 / RG-L)}& 32.99 / \textbf{14.83} / 30.84 & \textbf{33.23} / 14.77 / \textbf{31.00}\\
5&CNN/DM 10k {\fontsize{7pt}{0}\selectfont (RG-1 / RG-2 / RG-L)}& 39.34 / 17.30 / 36.75 & \textbf{39.71} / \textbf{17.55} / \textbf{37.18}\\
6&SQuAD QG {\fontsize{7pt}{0}\selectfont (Bleu-4 / MTR / RG-L)}& 23.23 / 25.47 / 51.25  & \textbf{23.34} / \textbf{25.54} / \textbf{51.30}\\
\bottomrule[1.0pt]	
\end{tabular}
}
\caption{Results of models pre-trained with typical generation and infilling generation. Tasks in the upper block are fine-tuned without noising, while the others are fine-tuned with noise-aware generation.}
\label{tab_8}
\end{table}
In Table \ref{tab_8}, we compare two ERNIE-G{\footnotesize EN}$_{BASE}$ variants that are pre-trained with typical generation mechanism and infilling generation mechanism and that generate word by word. Row 1-3 shows that without noising groundtruth texts, infilling generation outperforms typical generation across tasks. Furthermore, both variants achieve remarkable improvements by fine-tuning with noise-aware generation method (row 4-6). Specifically, Figure \ref{fig_4}a shows the results with diverse choices of noising rate $\rho_{f}$ on two tasks, indicating that appropriate noising substantially benefits the training and alleviates the training-inference discrepancy. 
To further analyze the excellence of infilling generation mechanism with noising, we compute the average attention weights of source tokens, unnoised target tokens and noised target tokens in the last self-attention layer respectively on 1,000 samples. Average attention weights with diverse noising rate $\rho_{f}$ are shown in Figure \ref{fig_4}b, which tells us that the model pays more attention on the decoder side to figure out noised points and assign them less attention weights as the noising rate $\rho_{f}$ increased in fine-tuning. Thereby, the model is able to detect and ignore the false predictions properly to alleviate accumulating mistakes while inference.

In column 1 of Table \ref{tab_7}, we compare four base size variants on three tasks. We see that noise-aware generation method and span-by-span generation task (rows 2-3 of Table \ref{tab_7}) play an important role in ERNIE-G{\footnotesize EN} pre-training and significantly outperform the baseline model which is only pre-trained with word-by-word infilling generation flow (row 4 of Table \ref{tab_7}). After integrating noise-aware generation method and span-by-span generation task, ERNIE-G{\footnotesize EN} boosts the performance across all three tasks, as shown in row 1 of Table \ref{tab_7}. In addition, U{\scriptsize NI}LM  is fine-tuned by masking words in the encoder and decoder to predict, which is also a case of noising for generation. To verify the idea that fine-tuning with masking language modeling like U{\scriptsize NI}LM is inefficient due to the coupling of masking (noising) and predicting that only the masked (noised) position will be learned, we also list the fine-tuning results obtained by predicting masked words with masking probability of 0.7, as shown in column 2 of Table \ref{tab_7}. We observe that our noise-aware generation method significantly outperforms the mask language modeling in seq2seq fine-tuning by predicting all words in the decoder side. 

\section{Conclusions}
 We present an enhanced multi-flow seq2seq pre-training and fine-tuning framework named ERNIE-G{\footnotesize EN} for language generation, which incorporates an infilling generation mechanism and a noise-aware generation method to alleviate the exposure bias. Besides, ERNIE-G{\footnotesize EN} integrates a new span-by-span generation task to train the model to generate texts like human writing, which further improves the performance on downstream tasks. Through  extensive experiments, ERNIE-G{\footnotesize EN} achieves state-of-the-art results on a range of NLG tasks. Future work includes incorporating reinforcement learning into pre-training for exposure bias and applying ERNIE-GEN to more NLG tasks such as machine translation.
 
\section*{Acknowledgments}
This work was supported by the National Key Research and Development Project of China (No. 2018AAA0101900).
\bibliographystyle{named}
\bibliography{ijcai20}

\begin{appendices}
  \section{Appendix}
  \label{app}
  \subsection{Pre-training on Large-scale Text Corpora}
  Recent works for pre-training verify that larger scaled pre-training corpora can improve the performances on downstream tasks. We pre-train ERNIE-G{\footnotesize EN}$_{LARGE}$ model on the 430GB text corpora with 1 epoch and 1M training steps. Our 430GB text corpora is extracted from the corpus used by RoBERTa \cite{roberta}, T5 \cite{T5} and ALBERT \cite{albert}. We fine-tune ERNIE-G{\footnotesize EN}$_{LARGE}$ on two abstractive summarization datasets including Gigaword and CNN/Daily Mail, the evaluation results are reported in Table \ref{tab_9}. Notice that the performance increase significantly as ERNIE-G{\footnotesize EN}$_{LARGE}$ pre-trains on larger scaled text corpora.
\begin{table}[H]
    \centering
    \setlength{\abovecaptionskip}{5pt}
    \setlength{\belowcaptionskip}{-0.3cm}
    \addtolength{\tabcolsep}{-1mm}
    \resizebox{0.485\textwidth}{!}{
    \begin{tabular}{l|cc|c}
    \toprule[1.0pt]
    \textbf{Model} & \textbf{{\footnotesize Data}} & \textbf{{\footnotesize Params}} & \textbf{{\footnotesize RG-1 / RG-2 / RG-L}}  \\
    \midrule[0.3pt]
    \multicolumn{4}{l}{\textbf{*Gigaword} \textit{10k}} \\
    ERNIE-G{\footnotesize EN}$_{LARGE}$&16G&340M&35.05 / 16.10 / 32.50\\
    \textbf{ERNIE-G{\footnotesize EN}}$\ddagger_{LARGE}$&430G&340M&\textbf{35.51} /
    \textbf{16.79} / \textbf{33.23}\\		
    \midrule[0.3pt]
    \multicolumn{4}{l}{\textbf{*Gigaword}} \\
    PEGASUS{\scriptsize (C4)} \cite{pegasus}&750G&568M&38.75 / 19.96 / 36.14\\
    PEGASUS{\scriptsize (HugeNews)} \cite{pegasus}&3.8T&568M&39.12 / 19.86 / 36.24\\
    ERNIE-G{\footnotesize EN}$_{LARGE}$&16G&340M&39.25 / 20.25 / 36.53\\
    \textbf{ERNIE-G{\footnotesize EN}}$\ddagger_{LARGE}$&430G&340M&\textbf{39.46} / \textbf{20.34} / \textbf{36.74}\\
    \midrule[0.3pt]
    \midrule[0.3pt]
    \multicolumn{4}{l}{\textbf{*CNN/Daily Mail}} \\
    T5$_{LARGE}$ \cite{T5}&750G&340M&42.50 / 20.68 / 39.75\\
    T5$_{XLARGE}$ \cite{T5}&750G&11B&43.52 / \textbf{21.55} / 40.69\\
    BART$_{LARGE}$ \cite{BART}&160G&400M&44.16 / 21.28 / 40.90\\
    PEGASUS{\scriptsize (C4)} \cite{pegasus}&750G&568M&43.90 / 21.20 / 40.76\\
    PEGASUS{\scriptsize (HugeNews)} \cite{pegasus}&3.8T&568M&44.17 / 21.47 / 41.11\\
    ERNIE-G{\footnotesize EN}$_{LARGE}$&16G&340M&44.02 / 21.17 / 41.26\\
    \textbf{ERNIE-G{\footnotesize EN}}$\ddagger_{LARGE}$&430G&340M&\textbf{44.31} / 21.35 / \textbf{41.60}\\
    \bottomrule[1.0pt]	
    \end{tabular}
    }
    \caption{Evaluation results on Gigaword and CNN/DailyMail for pre-trained models using large-scale text corpora. ERNIE-G{\fontsize{8pt}{0}\selectfont EN} with the $\ddagger$ mark is pre-trained with the 430GB text corpora.}
    \label{tab_9}
\end{table}
We also fine-tune ERNIE-G{\footnotesize EN} on the SQuAD 1.1 dataset for question generation task, the results are presented in Table \ref{tab_10}. We observe that larger scaled pre-training corpora can slightly improve the Rouge-L score and BLEU-4 score for the SQuAD dataset.
\begin{table}[H]
\centering
\setlength{\abovecaptionskip}{8pt}
\addtolength{\tabcolsep}{-1mm}
\resizebox{0.45\textwidth}{!}{
\begin{tabular}{l|cc|ccc}  
\toprule[1.0pt]
\textbf{Model} &\textbf{Data} &\textbf{Param} &\textbf{{\footnotesize BLEU-4}}~ & \textbf{{\footnotesize METEOR}} & \textbf{{\footnotesize ROUGE-L}}  \\
\midrule[0.3pt]
ERNIE-G{\footnotesize EN}$_{LARGE}$&16G&340M&\textbf{25.40}&\textbf{26.92}&52.84\\
\textbf{ERNIE-G{\footnotesize EN}}$\ddagger_{LARGE}$&430G&340M&\textbf{25.41}&52.84&\textbf{25.91}\\
\midrule[0.3pt]
\multicolumn{4}{l}{* \textit{Reversed test} $\leftrightarrow$ \textit{dev split}} \\
ERNIE-G{\footnotesize EN}$_{LARGE}$&16G&340M&26.95&\textbf{27.57}&53.77\\
\textbf{ERNIE-G{\footnotesize EN}}$\ddagger_{LARGE}$&430G&340M&\textbf{27.05}&27.43&\textbf{53.83}\\
\bottomrule[1.0pt]	
\end{tabular}
}
\caption{Question generation results on SQuAD. ERNIE-G{\fontsize{8pt}{0}\selectfont EN} with the $\ddagger$ mark is pre-trained with the 430GB text corpora.}
\label{tab_10}
\end{table}
The fine-tuning hyperparameters of ERNIE-G{\footnotesize EN}$\ddagger_{LARGE}$ are presented in Table \ref{tab_11}.
\begin{table}[H]
\centering
\setlength{\abovecaptionskip}{8pt}
\addtolength{\tabcolsep}{-1mm}
\resizebox{0.4\textwidth}{!}{
\begin{tabular}{l|ccc}  
\toprule[1.0pt]
 &\textbf{CNN/DailyMail} &\textbf{Gigaword} &\textbf{SQuAD QG}\\
\midrule[0.3pt]
Learning rate&4e-5&3e-5&1.25e-5\\
Batch size&32&128&32\\
Epochs&17&5&10\\
Dropout rate&0.1&0.2&0.2\\
Warmup ratio&0.02&0.1&0.25\\
Noising rate $\rho_{f}$&0.7&0.6&0.7\\
Label smooth&0.1&0.1&0.1\\
Beam size&5&6&5\\
Length penalty&1.2&0.7&1.0\\
\bottomrule[1.0pt]	
\end{tabular}
}
\caption{Hyperparameters used for fine-tuning on CNN/DailyMail, Gigaword, and SQuAD QG.}
\label{tab_11}
\end{table}
\end{appendices}

\end{document}